\documentclass[10pt,letterpaper]{article}
\usepackage[top=0.85in,left=2.75in,footskip=0.75in]{geometry}

\usepackage{amsmath,amssymb}

\usepackage{changepage}

\usepackage[utf8x]{inputenc}

\usepackage{textcomp,marvosym}

\usepackage{cite}

\usepackage{nameref,hyperref}

\usepackage[right]{lineno}

\usepackage{microtype}
\DisableLigatures[f]{encoding = *, family = * }

\usepackage[table]{xcolor}

\usepackage{fancyvrb}

\usepackage{array}

\usepackage{algorithm} 
\usepackage{algpseudocode} 
\usepackage{ulem}

\newcolumntype{+}{!{\vrule width 2pt}}

\newlength\savedwidth

\raggedright
\usepackage[aboveskip=1pt,labelfont=bf,labelsep=period,justification=raggedright,singlelinecheck=off]{caption}

\usepackage{comment}
\usepackage{todonotes}

\usepackage{booktabs}
\usepackage{multirow}

\makeatletter
\renewcommand{\@biblabel}[1]{\quad#1.}
\makeatother

\usepackage{lastpage,fancyhdr,graphicx}
\usepackage{epstopdf}
\fancyheadoffset[L]{2.25in}
\fancyfootoffset[L]{2.25in}
\begin{document}
\vspace*{0.2in}

\begin{flushleft}
{\Large
\textbf\newline{Calculating the matrix profile from noisy data} 
}
\newline
\\
Colin Hehir\textsuperscript{1},
Alan F. Smeaton\textsuperscript{1,2,*}
\\
\bigskip
\textbf{$^1$} School of Computing, Dublin City University, Glasnevin, Dublin 9, Ireland.\\

\textbf{$^2$} Insight Centre for Data Analytics, Dublin City University, Glasnevin, Dublin 9, Ireland.
\\
\bigskip

* alan.smeaton@DCU.ie

\end{flushleft}
\section*{Abstract}

The matrix profile (MP) is a data structure computed from a time series which encodes the data required to locate motifs and discords, corresponding to recurring patterns and outliers respectively. When the time series contains noisy data then the conventional approach is to pre-filter it in order to remove noise but this cannot apply in unsupervised settings where patterns and outliers are not annotated. The resilience of the algorithm used to generate the MP when faced with noisy data remains unknown.  
We measure the similarities between the MP from original time series data with  MPs generated from the same data with noisy data added under a range of parameter settings including adding duplicates and adding irrelevant data. We use three real world data sets drawn from diverse domains  for these experiments
Based on dissimilarities between the MPs, our results suggest that MP generation is resilient to a small amount of noise being introduced into the data but as the amount of noise increases this resilience disappears.

\section{Introduction}

Two well-used time series data mining examinations relate to motifs and discords \cite{mueen2009exact}. A time series motif is a pair of previously unknown sequences in a time series  or sub-sequences of a longer time series which are very similar to each other \cite{10.1007/978-3-642-34630-9_8} while a time series discord is  a sub-sequence of a long time series which is the most different from all the rest of the time series sub-sequences \cite{10.1007/978-3-030-28163-2_12}.
The matrix profile (MP) \cite{yeh2016matrix}, is a  data structure computed from a time series which locates the distance to, as well as the location of, the nearest neighbour of every sub-sequence in a time series. 
The MP  encodes all  details required to provide a solution for the detection of motifs and discords. This makes the MP suitable for detecting both outliers and recurring patterns. While detecting these can be regarded as classical AI problems and can be identified using other methods including machine learning, those other methods suffer from the curse of dimensionality or are complex and have multiple parameters to be adjusted \cite{luo2012efficient}. 
Anomalies and patterns are  important characteristics often studied in time series analysis. A similarity join \cite{10.1145/2000824.2000825} is a common technique for detecting such anomalies and patterns in a time series however it is inefficient whereas  MP algorithms can significantly reduce computing time for these tasks  \cite{zhu2016matrix}.

The MP offers a solution to detection of outliers and recurring patterns through efficient computation  while it is also able to consider sub-sequences of any length \textcolor{black}{and it has several advantages. These include that it is an exact solution and } 
provides no false positives or false negatives and it includes an exact solution for motif discovery, discord discovery, time series joins, etc.  \textcolor{black}{In contrast to other algorithms which require building and tuning access methods  the matrix profile is parameter-free and is space-efficient, with a space requirement which is linear in time series length with a small constant allowing processing of massively large data sets.  MP can also leverage parallel hardware including multicore processors and GPUs \cite{zhu2016matrix}.  It is domain agnostic and requires only one input parameter, the sub-sequence length $m$.  It has a time complexity of $O(n^2\log{(n)})$ that is constant across sub-sequence lengths \cite{yeh2016matrix}.  It can be re-computed incrementally as a time series grows and thus it can support real time applications \cite{yeh2016matrix}.  A variation of the matrix profile algorithm called the Motif Discovery with Missing Data (MDMS) has recently been introduced \cite{8888196} with the ability to handle missing data in that it can provide answers guaranteed to have no false negatives but which may have false positives and we shall return to this point later.}

The two main components of the matrix profile are a distance profile and a profile index. A vector of minimum Z-Normalised Euclidean Distances constitutes the distance profile. The initial nearest-neighbour index is included in the profile index which is essentially the position of the sequence's most comparable sub-sequence \cite{yeh2016matrix}.
\textcolor{black}{
In summary, the steps for computing the matrix profile from a time series $X$ are as follows:
\begin{enumerate}
    \item Choose a subsequence length $m$ that is appropriate for the application. This length should be smaller than the length of the time series $X$ that is being analysed;
\item Compute the matrix profile, which is an array of length $n-m+1$ where $n$ is the length of the time series $X$. The matrix profile contains the distances between each subsequence of length $m$ in $X$ and its nearest neighbour subsequence in $X$.
\item Compute the matrix profile index, which is an array of length $n-m+1$ that contains  indices of the nearest neighbour subsequence for each subsequence in $X$. This can be used to quickly retrieve  nearest neighbour subsequences.
\end{enumerate}
}

After computing the matrix profile and matrix profile index, they can be used to efficiently perform various time series data mining tasks including exact motif discovery, anomaly detection, and similarity search.

\begin{figure}[!ht]
\centering 
\includegraphics[width=\columnwidth]{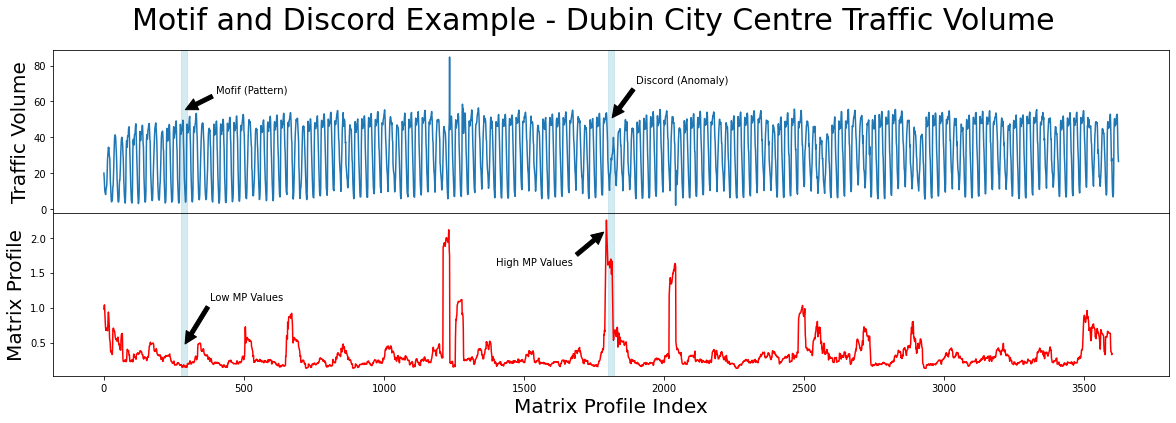}
         \caption{Sample \textcolor{black}{time series} data from Dublin city centre traffic \textcolor{black}{flow} with matrix profile illustrating motif and discord regions. \label{fig:MP-D}}
\end{figure}

Fig~\ref{fig:MP-D}  shows an example of some original time series data representing the volume of traffic in Dublin City Centre and the MP plot derived from that data. The motif is a repeated pattern in the original time series with a matching area demonstrating low MP distance values while the discord or anomaly is a mismatch region demonstrating high MP distance values. Even if the MP distance value evaluated is non-zero, a localised MP minimum value may be utilised to detect a near match, which is an essential characteristic of the MP \cite{yeh2016matrix}.

There are numerous algorithms that compute the MP. The brute force approach of the Naive algorithm is inefficient and the current best-in-class is  {SCRIMP++}  \cite{zhu2018matrix}. 
STAMP Incremental, or STAMPI, is also useful because it facilities the MP to be incrementally maintained \cite{yeh2016matrix} which means it can be used in real time applications.
Once a MP is generated there are techniques to extract the top-K  repeated patterns/anomalies in a given time series using the MP data structure \cite{yeh2018time} and to perform segmentation analysis to allow navigation through the resultant MP.

Applications of the MP for time series data mining have already generated many insights \cite{zhu2020swiss}.
One study discovered motifs using MP in retail product sales time series and used them to analyse the temporary sales correlations among products thus indicating that customers’ product preferences are not stable and change with time \cite{li2021temporary}. The MP has also been used in anomaly detection on IT operations time series data to address the issue of monitoring IT systems' Key Performance Indicators \cite{lan2021anomaly}.  The MP has  offered market analysis based techniques in terms of stock-market financial time series data \cite{cartwright2021financial}, while recent research has shown that in predicting COVID-19 cases, a hybrid of the MP and an attention-based long short-term memory (LSTM) model performed best when compared to other  models \cite{liu2021novel}.

The MP algorithm has enabled the discovery of motifs from time series of substantial lengths  where previously the memory and processing requirements obstructed exact motif search from time series which have a length of more than one hundred million data points \cite{zhu2016matrix}.
This scalability characteristic of the MP algorithm is  attractive but its most useful and important feature is its generalisation across any application-agnostic time series \cite{dau2017matrix}. 

One of the known drawbacks with the MP is its performance on large-scale sets of noisy data, such as occur in most natural applications \cite{yeh2018time} and that is \textcolor{black}{the specific issue} we focus on in this paper \textcolor{black}{and where we make our contribution}.
Many time series data in real world applications have noise which could interfere with the generation of an accurate MP.  In some forms of time series analysis such as generating periodograms, algorithms such as the Lomb-Scargle \cite{doi:10.1076/brhm.30.2.178.1422} have been developed
that are tolerant to unequally sampled data, to data sets with missing values and to \textcolor{black}{data sets with} other forms of noise.  In such cases it is the tolerance of the algorithm itself that handles the noise in the data \textcolor{black}{but that is not the case for algorithms which generate the MP}.

\textcolor{black}{Earlier we mentioned that a very recent variation of the matrix profile algorithm called MDMS can handle missing data in the original time series by providing answers which are guaranteed to have no false negatives \cite{8888196}. That paper acknowledges that there is no other algorithm that can find motifs in the way the matrix profile does, in the presence of missing data. The paper  generates pseudo missing data in the same way as we do here, and their definition of missing data covers random insertions (referred to as ``spikes'', noise or corrupted data and gaps in data capture corresponding to blocks of missing data. The work uses data from two case studies, seismological data and activity data, but the amount of data corruption is quite small, corresponding to deletion of 50 individual data points and removal of two blocks of data of length 25 values each from time series of the order of thousands of values.  The performance of the MDMS algorithm is evaluated by examining the before and after matrix profiles and comparing the resulting graphs manually rather than in a quantitative way.}

\textcolor{black}{Prior to the very recent single example of work on developing an algorithm to handle missing or noisy data in generating the matrix profile mentioned above, the standard approach has been to eliminate the noise from, or to fill the gaps in the data. A recent example of that approach can be seen in work by Berjab {\it et al.} \cite{9792239} where the authors concentrated on recovering missing data but do not deal with other forms of noisy data. That work  focused on detecting false data injection attacks with missing data appearing with probabilities between 0.001 and 0.002 in two test datasets of 2.5 million and 20.9 million data points respectively. Our work here focuses on real world cases where noisy data can take many other forms and can occur much more frequently.}

De Paepe {\it et al.}
\cite{10.1007/978-3-030-40014-9_5} have recently applied  noise elimination  on real internet traffic time series data and subsequently detected anomalous behaviours through generating a matrix profile. Related work in  \cite{icpram19} has demonstrated how the elimination of noise as a pre-process to MP generation can help in anomaly detection from noisy data. This was tested on the Numenta Anomoly Benchmark \cite{7424283}, a well-known collection of data sets which focus on detecting anomalies from time series  which contain noisy data.  The Numenta Anomoly Benchmark  has recently been superseded by the more comprehensive ADBench \cite{han2022adbench} which has 57 data sets each with different noise levels and it benchmarks the effectiveness of 30 different algorithms for anomaly detection on noisy data. The noise filtering in  \cite{icpram19} was achieved with the same overall computational complexity as MP generation but  was tested on a time series of only 2,000 data values with synthetic noise added. Furthermore there was no investigation into the impact different amounts of noise have on the generation of the MP. It may be that MP generation has a tolerance to a certain amount of noise inherent in the data used to generate it that we do not yet know about \textcolor{black}{and that is what we investigate and report on here}.

In the work in this paper we  measure the effect of noise on  standard MP generation without the overhead of pre-filtering as reported in \cite{10.1007/978-3-030-40014-9_5,9792239}  and elsewhere. Our motivation is that pre-filtering noise from a time series may also dilute whatever anomalies, discords or motifs exist in the original data. In this paper we generate MPs from three data sets of different sizes and  we artificially introduce  noise at different levels of intensity to \textcolor{black}{\sout{pollute}corrupt} each data set using  noise creation techniques and parameters from ADBench \cite{han2022adbench}. \textcolor{black}{The amount and types of data corruption we introduce into the original datasets are far in excess of those reported in other work which uses either noise elimination or works with missing data \cite{8888196}.}
We then re-generate MPs and compare the characteristics of MPs on clean data with the equivalent MPs from data with noise added \textcolor{black}{where the amount and types of noise introduced to corrupt the data is more realistic than reported elsewhere, and that is the main contribution of this paper}. This addresses the underlying research question of what is the actual impact of noise \textcolor{black}{of different types and different magnitudes,} on MP outputs.

\section{Materials and methods}

We present details of three case studies where we apply the matrix profile in different domains to time series data from real-world scenarios and then  we describe how we add  noise to each data set.

\subsection{Case study 1: keystroke timings}

Lifelogging involves gathering digital records or logs of a person's lifestyle, activities, and encounters during a typical day,  in an automatic fashion \cite{gurrin2014lifelogging}. Such data is collected by an individual and is not normally shared with others or made public. Lifelogs represent a personal record that may  be analysed either directly by the individual gathering the data, or by others on their behalf \cite{tuovinen2022privacy}. This is done in order to observe long-term behavioural patterns and changes in terms of health, well-being or cognitive changes, as well as to facilitate  retrieval of information from the individual's past \cite{llogapplics}. 

One form of automatic lifelogging is keystroke dynamics which uses a software application, a keystroke logger, to collect timing information about every key pressed on a keyboard or mobile device when the individual has been typing \cite{smeaton2021keystroke}.
Precise timing information on keystrokes, namely the time taken to type  two  adjacent characters, is captured to the nearest millisecond. We can then examine inter-keystroke timings to determine  differences among individuals or differences within the timings for an individual. The initial application of this  analysis was in the area of  user authentication based on the feature that every individual has unique keystroke timing patterns \cite{joyce1990identity}. A more recent application is  analysing a user's cognitive processes while typing, where we compare timings from an individual's baseline gathered over a period of time  with the current dynamic  \cite{leijten2013keystroke} to determine writing fluency.  In turn,  writing fluency can reveal when the subject was pausing and revising their writing indicating revision rather than creation of new material. 

Keystroke timing data was collected in a previous study using the Loggerman logging tool \cite{hinbarji2016loggerman}. Timing information for keystrokes was obtained for one  user for 2,522,186 characters typed over a 12-month period and the data is available at \cite{Smeaton2020}.  For  privacy reasons the specific characters typed were anonymised via random mapping which is consistent across the  data-set, permitting the extraction of character and bigram timing information.  This typing data is already noisy because of occasional data missing because  the Loggerman tool stops recording keystrokes when it suspects the user is about to enter a username, password or other confidential information and it does this in a very conservative way.  The keystroke timing information  in \cite{Smeaton2020} was processed to compute the time elapsed between all adjacent typed characters and those bigrams typed greater than 1,000ms apart were removed. In this paper we focus on the timing for most frequently occurring bigram which occurs 56,545 times when typed in less than 1,000ms during the several months of recording. 

\begin{figure}[!ht]
\centering
\includegraphics[width=\columnwidth]{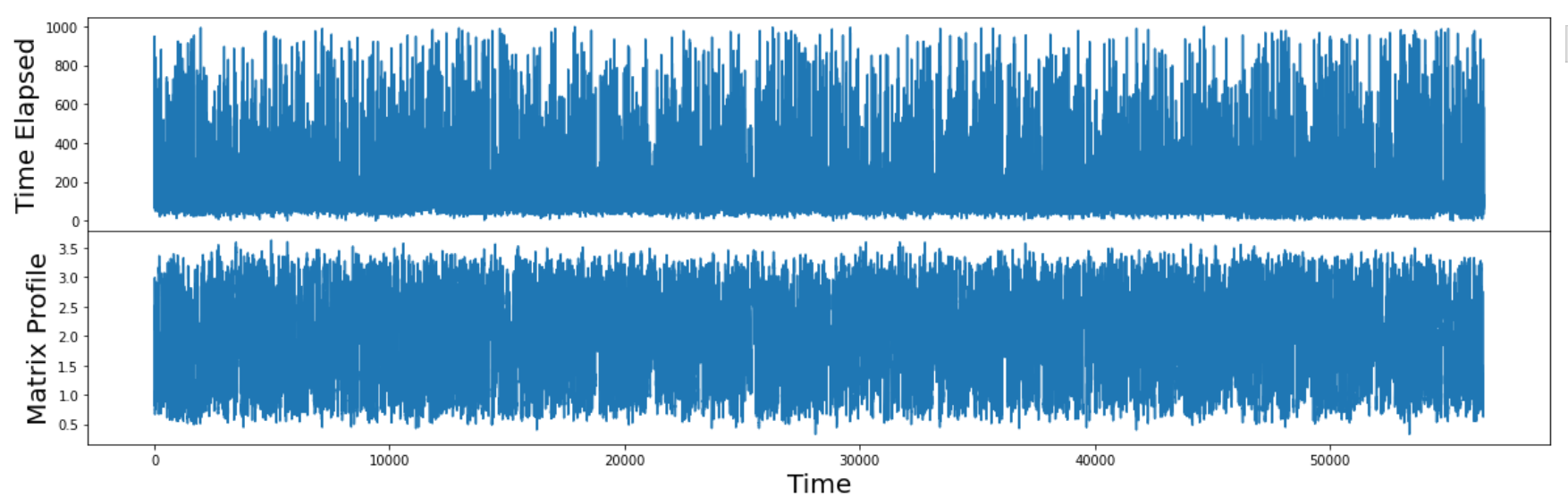}
   \caption{Occurrences and MP for the top occurring bigram \label{fig:bimp1}} 
\end{figure}

To calculate the MP on the keystroke timing data the only parameter needed  is the window size, in addition to the time series of 56,545 values. Our choice of  window size was a sequence of 20 characters, large enough to capture   patterns, while  not being smaller than a potential sub-sequence pattern. Once generated, the matrix profile provides an array of z-normalised Euclidean distances to their nearest neighbour (i.e. the MP values), including other values such as the MP indices. 
Fig~\ref{fig:bimp1} 
shows the 56,545 keystroke timing occurrences and the MP for the most frequently occurring bigram. 

\subsection{Case study 2: movement sensors on new-born calves}

Improving efficiency in the area of animal management and livestock welfare  has resulted in the emergence and use of precision agriculture technologies \cite{shalloo2021review}. This includes the gathering of continuous data on the activities and behaviour of cattle which has great potential for both effective food production and improved animal welfare \cite{hut2022sensor}.  

Wearable 3-D accelerometers can be used to monitor animal behaviour \cite{riaboff2022predicting}. 
In the case of new born calves, an accelerometer attached to a collar around the neck was used to measure movements such as walking, trotting and running. Locomotor play, a repetitive and exaggerated movement, is also demonstrable in young calves via behaviour consisting of jumping, bucking and running  \cite{riaboff2022predicting}. 

Raw data from a neck-worn AX3 accelerometer sensor \cite{doherty2017large} on  new-born calves was used where the sensor was worn from birth for several weeks by each calf.  The data is available at \cite{Smeaton2021}. 
Attributes from the movement data include the timestamp and the values for x-acceleration, y-acceleration, and z-acceleration. An additional movement attribute independent of sensor orientation was derived known as the acceleration magnitude  ($A_{mag}$), for use as a single time-series for analysis \cite{riaboff2022predicting} and shown in Eq~\ref{eq:amag}. This  eliminates the impact of rotation of the sensor around its neck  of the calf as the collar rotates and  produces a measure of how quickly the velocity of the calf changes in any direction.  

\begin{eqnarray}
\label{eq:amag}
  A_{mag} = \sqrt{{accel_x}^2 + {accel_y}^2 + {accel_z}^2}
\end{eqnarray}

\noindent
The sampling frequency for  the AX3 accelerometer was 12.5Hz, resulting in more than one hundred million data points from each calf's accelerometer over the logging period which was approximately 6 weeks. Movement values were re-sampled to one minute intervals by calculating the mean of the $A_{mag}$ values contained within  non-overlapping one minute intervals. 
Similar to the  case study on keystroke dynamics, a window size was selected to generate a MP for the acceleration magnitude. The raw movement data aggregated to one sample  every minute for the logging period yielded 60,480 data points per calf and
a value of 60 was chosen for the window size when generating the MP to represent a span of one hour to capture any  recurring motifs and discords. For the purpose of generating the MP and adding noise, we used the movement data from one calf from the herd.

\subsection{Case study 3: city centre traffic volumes}

\begin{figure}[!htb]
\centering
\includegraphics[width=\columnwidth]{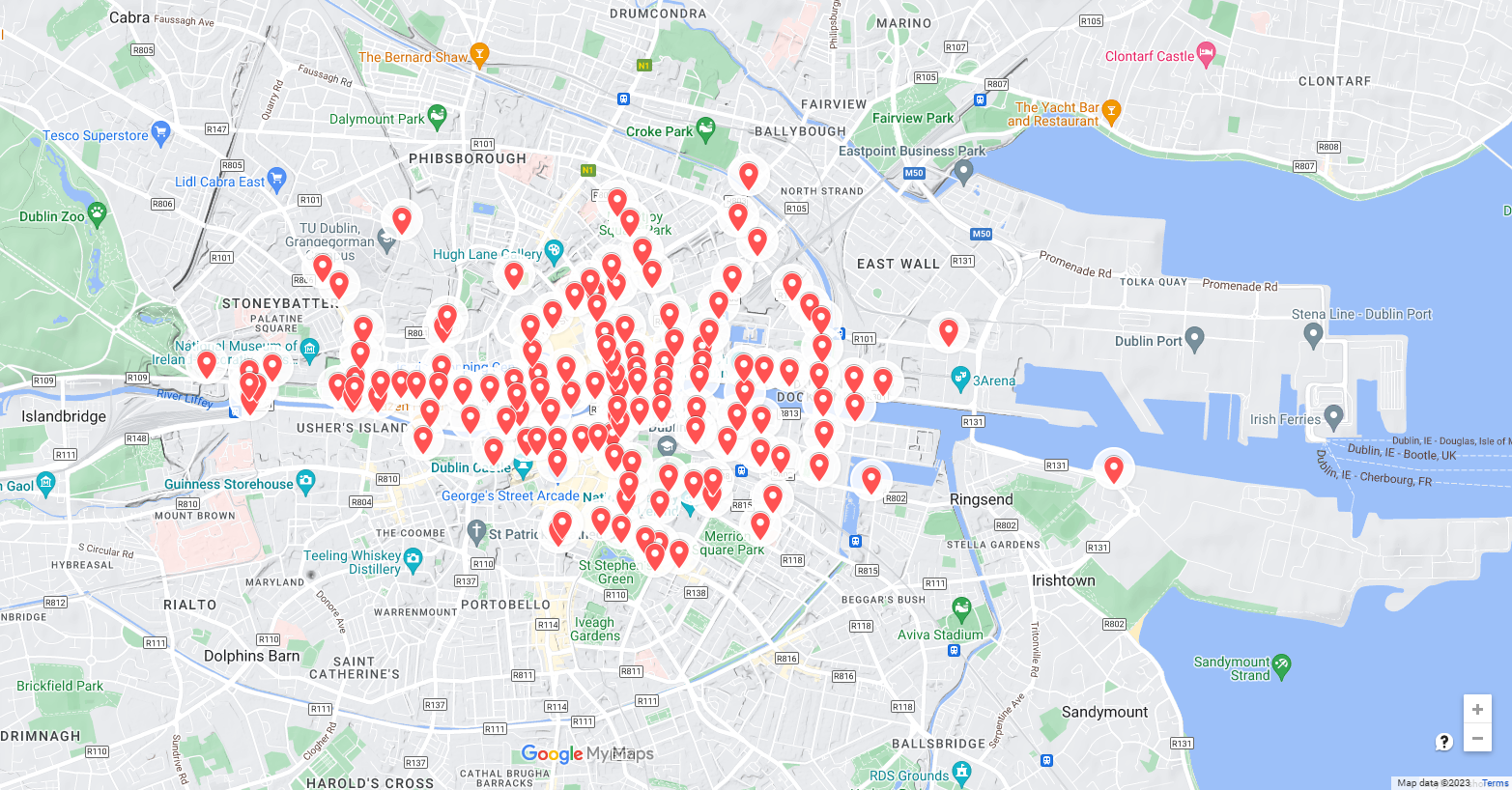}
\caption{Traffic signals and 132 SCATS  sensor locations in Dublin city centre marked as red dots.\label{fig:scats}}
\end{figure}

The  {\it Sydney Coordinated Adaptive Traffic System} (SCATS) is used to collect traffic volume data across many cities worldwide, including  Dublin, Ireland  \cite{mccann2014review}. By also managing the timing of traffic signals to control traffic flow, SCATS acts as an intelligent transportation system. The mechanics of its operation are that it detects vehicle presence in each lane at points on roads typically just before junctions,  as shown in Fig~\ref{fig:scats} as well as counting the number of pedestrians at sites waiting to cross a road. The sensors  are installed under the road surface as inductive loops and the data from all the sensors in a city feeds into a control system for the city's traffic management. While it has  advantages, the SCATS data is noisy  and it cannot provide insights into what is normal traffic behaviour or what are deviations from that normal behaviour. This results in  reactive responses by control room staff who monitor data streams  manually \cite{kinane2014intelligent}. 
Since a labelled training data-set is not  available  it is not possible to detect traffic anomalies by building a classification model.

SCATS traffic volume data from January to May 2022 was provided for analysis by Dublin City Council and is available at \url{https://data.gov.ie/dataset/dcc-scats-detector-volume-jan-jun-2022}. A count of the volume of traffic  on approaches to road junctions is an indicative representation of overall traffic flow within  regions of the city. 
The accuracy of data from the sensors at  each site  cannot be guaranteed due to  faulty detectors or to sensor communication issues thus making this data noisy. In order to  indicate patterns in the flow of traffic for the city centre as well as to combat data collection errors, the sum of total traffic volume for the city centre was computed per hour for a 5-month logging period. A window size of 24 was selected to generate the MP, representing a full one day. The total amount of data consisted of 5 months recording from 132 traffic sensors with overall traffic volume sampled hourly yielding a total of 475,200 individual data values aggregated into a time series of 3,600 data points.  

\begin{figure}[htb]
\centering
\includegraphics[width=\columnwidth]{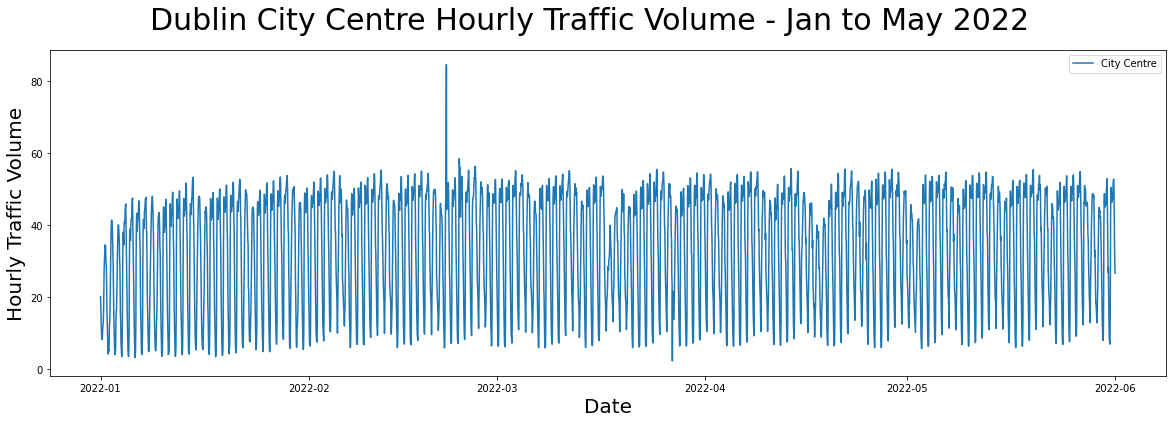}
\caption{Traffic volume  for Dublin city between January and May 2022. \label{fig:traffic-vol}}
\end{figure}

\begin{figure}[htb]
\centering
\includegraphics[width=\columnwidth]{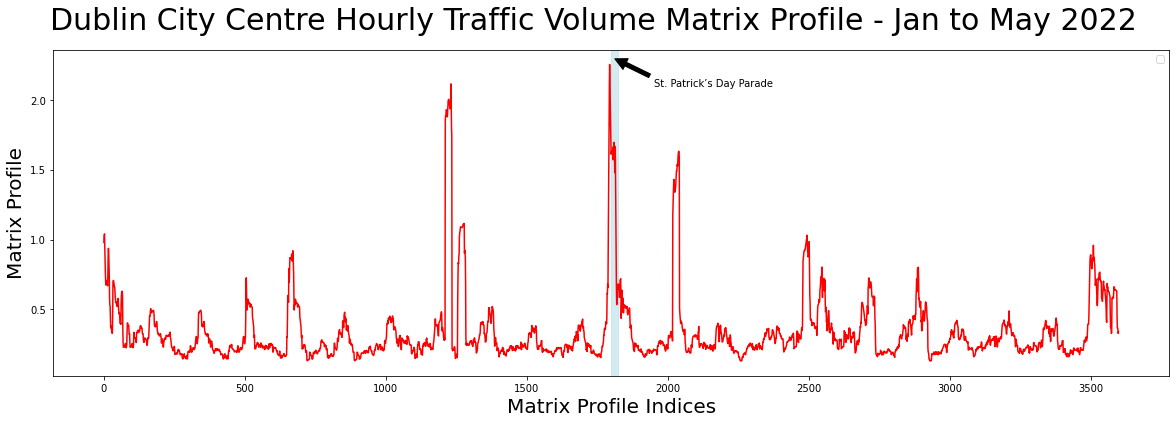}
\caption{MP values  for Dublin city between January and May 2022. \label{fig:traffic-MP}}
\end{figure}

Fig~\ref{fig:traffic-vol} shows the time series for the hourly traffic volume for Dublin city showing a regular daily pattern while Fig~\ref{fig:traffic-MP} shows the matrix profile for the traffic volume with some discords and motifs.

\subsection{Adding noise to time series data}

Since one of the  objectives of the matrix profile is to  pinpoint instances in a time series that deviate significantly from the time-series as a whole, there is interest to determine  the limitations to the MP in terms of  stability and robustness under different levels of data  noise.  As noted in \cite{10.1007/978-3-030-40014-9_5} and  \cite{icpram19}, time-series data from real-world applications usually suffer from noise and data corruption to some extent. However the impact of this on the matrix profile has only been researched in relation to noise elimination as a pre-filter before generating a MP.  This paper examines the impact of noise on the MP algorithm without pre-filtering.

The ADBench \cite{han2022adbench} is a comprehensive review of 30 algorithms for anomaly detection on 57 benchmark data sets and covers their performances under different levels of supervision, anomaly types and noisy and corrupted data. For the  noise and data corruption settings, ADBench considers three  types of noise namely duplicated anomalies, the insertion of irrelevant features, and annotation errors. 
Interestingly, these anomalies all involve additions to the time series data and not the removal of any data but because that is what ADBench does, we will do likewise in this paper.  

In an ideal situation, to evaluate the tolerance of the MP to noisy data we would generate MPs from data sets with labelled anomalies in a supervised setting, we would add noise to these, recompute the MPs and compare the characteristics of the before and after MPs. Since the three data-sets used in this study are unlabelled we use workarounds similar to those who used the Numenta benchmark in \cite{7424283}.

ADBench defines  duplicate anomalies as  likely to repeat multiple times in  data for reasons such as recording errors. 
We added duplicated data values to each of our three data sets using the same parameters as  Adbench  namely that a randomly selected 5\% of additional data values were denoted as anomalies  then duplicated up to 6 times in multiple runs. That means up to an additional 25\% of  data added as we add duplicates times 2, times 3, times 4, times 5 and times 6.

For the irrelevant features noise type, ADBench indicates these are likely to be caused by measurement noise or inconsistent measuring units meaning that detecting anomalies such as discords and motifs in a time-series would be  more difficult as they may be hidden. 
We added irrelevant features to each of our three real-world data sets using the same parameters as in ADBench.
Irrelevant data points were randomly added  values to each time series in stages with additions of 1\%, 5\%, 10\%, 25\% and up to 50\% of the total data points. The added values came from generating  features from the uniform distribution 
\[Unif~(min(\mathbf{X}),max(\mathbf{X}))\]   
\textcolor{black}{where $min()$ and $max()$ refer to the minimum and maximum values in the time series $X$ and where we assume data values in the time series have a normal distribution.} We include these in the original data. Irrelevant data points were added randomly into the original series without shuffling the order of data points in the original series. For  adding  noise to our three data sets we used the code available at  \url{https://github.com/Minqi824/ADBench/}

In order to measure the impact of noise on the data used to generate a MP, we compare the MPs generated from clean data and from data with noise added.  In effect the comparison of two MPs  is the comparison between two resulting time series.  The metrics we use to describe and compare MPs include the mean, maximum and minimum values of each MP though we are wary that these descriptive statistics can hide much of the similarities or differences between MPs, as illustrated by  Anscombe's Quartet \cite{anscombe1973graphs}.

Since the two MPs being compared will be of different lengths because of the addition rather than  removal of noise data to the original, increasing it by up to 25\% of the total MP length in some cases, we use dynamic time warping \cite{vaughan2016comparing}  to account for matching under this constraint. The implementation we use is FastDTW developed by Salvador and Phillip \cite {salvador2007toward} and the distance measure between the two MPs is the absolute difference between matched values. This is a dissimilarity measure with identical MPs yielding a value of 0 and as the value increases it reflects increasingly dissimilar MPs.

\begin{figure}[htb]
\centering
\includegraphics[width=\columnwidth]{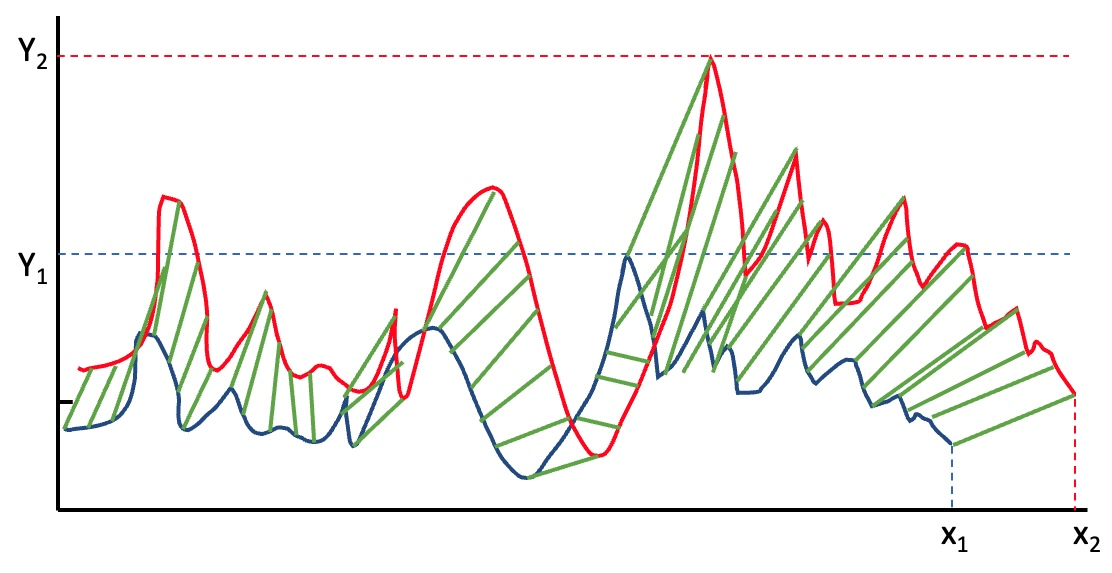}
\caption{Schematic of  DTW dissimilarity calculation and normalisation between MPs. \label{fig:DTW}}
\end{figure}

Figure~\ref{fig:DTW} shows a schematic of how MP dissimilarity is calculated for   data which has  noise from duplicated anomalies x2 added. The original MP has length $x_1$  and the duplicates add another 5\% making the length $x_2$. The MP for the original data is shown in blue, and the MP for the  data with noise is shown in red. The  lengths of the green lines (there should be one for each  data point in the original data) correspond to  absolute differences between corresponding values from the MPs as matched by the dynamic time warping algorithm. 
Each of the generated MPs will have a  maximum value which is $y_1$ in the case of the MP from the original data (in blue) and $y_2$ in the case of the MP from the noisy data. To normalise the  dissimilarity between the two MPs we divide the sum of the absolute differences between corresponding values (the sum of the lengths of the green lines in Fig~\ref{fig:DTW} by the the number of values in the original time series and also by the maximum value of the MP from the original data, $y_1$.

In FastDTW the radius parameter is  used to approximate the exact DTW. If the radius is equal to the length of the times series being examined,  then FastDTW is optimised and is  equal to DTW. If the radius is less the time series length then FastDTW is not as accurate as DTW but  is more computationally efficient though this has been questioned recently \cite{wu2020fastdtw}.
We selected a radius of 30 for FastDTW for all  data as a fixed width radius is sufficient so long as it is ``wide enough" to allow the insertion of  duplicates and irrelevant features without disruption of the similarity computation.

We now present the results from generating MPs on each of the original time series data sets and on MPs where noise has been added. For these time series  there is good variety among the sizes of the time series with sizes of 3,600 (traffic), 56,545 (keystrokes) and 60,480 (calf movements).

\section{Results}

We added noise to the original  data sets consisting of duplicated anomalies up to 6 times and irrelevant features up to 50\%,  regenerated MPs for the 10 different parameter settings and in  Tables~\ref{tab:keystroke_anomaly_results}, \ref{tab:calf_anomaly_results} and \ref{tab:traffic_anomaly_results} we compare the MPs from noisy data against the MP for the original data, for each data set.  

\begin{table}[!ht]
\centering
\begin{tabular}{c|l|c|c|c|c} 
 \toprule
 \textbf{Signal} & \textbf{Type} & {\bf $\Sigma$ abs diffs} &\textbf{Mean}  & \textbf{Max} & \textbf{Min}\\ 
 && {\bf  MP values}&&& \\
 \midrule
 \multirow{6}{*}{Keystrokes} & Original Matrix Profile &  0 & 1.87 & 3.63 & 0.33\\
  & Duplicated Anomaly $\times$ 2 &  {\bf 9,306} & 1.88 & 3.71 & 0.37\\
  & Duplicated Anomaly $\times$ 3 &  10,669 & 1.88 & 3.58 & 0.36\\
  & Duplicated Anomaly $\times$ 4 &  11,856 & 1.86 & 3.61 & 0.34\\
  & Duplicated Anomaly $\times$ 5 &  12,820 & 1.86 & 3.58 & 0.39\\
  & Duplicated Anomaly $\times$ 6 &  14,053 & 1.86 & 3.56 & 0.39\\
\midrule
 \multirow{5}{*}{Keystrokes} & Irrelevant Features - 1\% &  9,683 & 1.80 & 3.64 & 0.30\\
  & Irrelevant Features - 5\% &  {\bf 15,333} & 1.54 & 3.76 & 0.33\\
  & Irrelevant Features - 10\% &  19,528 &  1.39 & 3.85 & 0.38\\
  & Irrelevant Features - 25\% &  25,787 & 1.42 & 3.86 & 0.42\\
  & Irrelevant Features - 50\% &  27,368 & 1.66 & 3.2 & 0.47\\
 \bottomrule
\end{tabular}
\caption{Matrix Profile value changes for keystroke timing data (N=56,545) with noisy data added under different parameter settings.\label{tab:keystroke_anomaly_results}}
\end{table}

\begin{table}[!ht]
\centering
\begin{tabular}{c|l|c|c|c|c} 
 \toprule
 \textbf{Signal} & \textbf{Type} & {\bf $\Sigma$ abs diffs}  & \textbf{Mean}  & \textbf{Max} & \textbf{Min}\\ 
 && {\bf  MP values}&&& \\[0.5ex] 
 \midrule
 \multirow{6}{*}{Calf $A_{Mag}$} & Original Matrix Profile & 0 & 5.10 & 8.28 & 0.63\\
 & Duplicated Anomaly $\times$ 2 & {\bf 23,585} & 5.41 & 8.05 & 0.06\\
 & Duplicated Anomaly $\times$ 3 & 31,467 & 5.71 & 8.15 & 0.09\\
 & Duplicated Anomaly $\times$ 4 & 38,424 & 5.93 & 8.21 & 2.32\\
  & Duplicated Anomaly $\times$ 5 & 50,321 & 6.04 & 8.21 & 2.79\\
 & Duplicated Anomaly $\times$ 6 & 51,830 & 6.19 & 8.27 & 2.45\\
\midrule
 \multirow{5}{*}{Calf $A_{Mag}$} & Irrelevant Features - 1\% & 19,811 & 5.09 & 8.09 & 0.05\\
 & Irrelevant Features - 5\% & {\bf 32,825} & 5.46 & 7.90 & 0.06\\
 & Irrelevant Features - 10\% & 47,781 & 5.92 & 7.85 & 2.10\\
 & Irrelevant Features - 25\% & 97,028 & 6.68 & 7.96 & 3.88\\
 & Irrelevant Features - 50\% & 117,820 & 7.01 & 8.13 & 5.12\\
 \bottomrule
\end{tabular}
\caption{Matrix Profile value changes for calf movement data (N=60,480)  with with noisy data added under different parameter settings.\label{tab:calf_anomaly_results}}
\end{table}

\begin{table}[!ht]
\centering
\begin{tabular}{c|l|c|c|c|c} 
 \toprule
 \textbf{Signal} & \textbf{Type} & {\bf $\Sigma$ abs diffs}  & \textbf{Mean}  & \textbf{Max} & \textbf{Min}\\
 && {\bf  MP values}&&& \\ [0.5ex] 
 \midrule
 \multirow{6}{*}{Traffic} & Original Matrix Profile & 0 & 0.34 & 2.25 & 0.13\\
 & Duplicated Anomaly $\times$ 2 & {\bf 1,739} & 0.91 & 4.09 & 0.13\\
 & Duplicated Anomaly $\times$ 3 & 2,724 & 1.39 & 4.23 & 0.20\\
 & Duplicated Anomaly $\times$ 4 & 2,992 & 1.70 & 4.49 & 0.20\\
 & Duplicated Anomaly $\times$ 5 & 3,864 & 1.99 & 4.52 & 0.26\\
 & Duplicated Anomaly $\times$ 6 & 4,701 & 2.17 & 4.66 & 0.48\\
\midrule
 \multirow{5}{*}{Traffic} & Irrelevant Features - 1\% & 764 & 0.54 & 3.29 & 0.13\\
 & Irrelevant Features - 5\% & {\bf 2,109} & 1.08 & 3.96 & 0.16\\
 & Irrelevant Features - 10\% & 2,705 & 1.59 & 3.84 & 0.20\\
 & Irrelevant Features - 25\% & 4,567 & 2.47 & 4.46 & 0.61\\
 & Irrelevant Features - 50\% & 11,617 & 3.17 & 4.46 & 1.34\\
 \bottomrule
\end{tabular}
\caption{Matrix Profile value changes for traffic movement data (N=3,600) with noisy data added under different parameter settings.\label{tab:traffic_anomaly_results}}
\end{table}

\textcolor{black}{Other work which has examined the impact of noise on the matrix profile has done so by directly comparing the two matrix profile graphs, one before and one after corrupting the time series data \cite{8888196}  In our work rather than ``eyeball'' the two MP graphs which would be unwieldy because of their sizes (56,545, 60,480 and 3,600 data points respectively) we extract quantitative characteristics of the before and after matrix profile graphs, and this is part of the novelty of this paper.}  In addition to presenting the mean, maximum and minimum values of the \textcolor{black}{before and after} matrix profiles, as well as the  the sum of the absolute differences between corresponding MP values, in Table~\ref{tab:similarity_results} we present the normalised dissimilarities for each noise parameter setting and for each data set.

\begin{table}[!htb]
\centering
\begin{tabular}{r|c|c|c} 
 \toprule
\textbf{Data set} &\textbf{~Keystrokes~}&	\textbf{~~Calf $A_{Mag}$~~}&	\textbf{~~Traffic~~}\\
N =  & 56,545 & 60,480 & 3,600 \\
\midrule
\textbf{MP Length}&	56,545	&60,480&	3,600\\
\textbf{Maximum value}	&3.63 &	8.28	&2.25\\
\midrule 
Duplicated Anomaly $\times$ 2	& 0.045	& 0.047	& 0.214\\
Duplicated Anomaly $\times$ 3	& {\bf 0.051}	& {\bf 0.062}	& {\bf 0.336}\\ 
Duplicated Anomaly $\times$ 4	& 0.057	& 0.076	& 0.369\\ 
Duplicated Anomaly $\times$ 5	& 0.062	& 0.100	& 0.477\\ 
Duplicated Anomaly $\times$ 6	& 0.068	& 0.103	& 0.580\\
\midrule			
Irrelevant Features - ~1\%	& 0.047	& 0.039	& 0.094\\
Irrelevant Features - ~5\%	& {\bf 0.074}	& {\bf 0.065}	& {\bf 0.260}\\
Irrelevant Features - 10\%	& 0.095	& 0.095	& 0.333\\
Irrelevant Features - 25\%	& 0.125	& 0.193	& 0.563\\
Irrelevant Features - 50\%	& 0.133	& 0.235	& 1.434\\
\bottomrule
\end{tabular}
\caption{Normalised dissimilarities between MPs generated from clean data and from data with various noise parameter settings, for each data set. \label{tab:similarity_results}}
\end{table}

\section{Discussion}

By introducing noise \textcolor{black}{into a time series of data}, which may or may not be anomalies, this will disrupt the generated MP and thus the noise itself will be detected as anomalies because the MP cannot make a distinction between noise and real data.  In particular, duplicates introduced as noise will appear in a MP as a pattern and will really disrupt the generated MP.

The results in Tables~\ref{tab:keystroke_anomaly_results}, \ref{tab:calf_anomaly_results} and \ref{tab:traffic_anomaly_results} as well as in Table~\ref{tab:similarity_results} present several insights which we discuss in turn. 
Before examining these we  point out that \textcolor{black}{at the high end of the data corruption} the amount of noise we introduce to the time series is exceptionally large and not likely to occur in any useful real world application. Our reason for going to these extreme noise levels is to discover the resilience of the MP generation algorithm at all noise levels, from minor to very large. Thus in examining the results,  the most important are at where we introduce duplicate anomalies x2 and irrelevant features at 1\% or 5\%.

The introduction of irrelevant features have greater impact on generated MPs than the introduction of duplicate values, making the MPs more dissimilar to the MP on the original data. This occurs for each data set and arises because this noise parameter  pollutes the original time series extensively, adding up to 50\% additional noisy data. We also observe that MPs for noisy traffic data (N=3,600) are more dissimilar to the MP from the original data than for the keystrokes data (N=56,545) which in turn are less than for the calf data (N=60,480).
This means that the longer the time series, the more dissimilar the resulting MPs according to FastDTW. This can be seen when comparing dissimilarity values across columns in Fig~\ref{tab:similarity_results} as well as in the sums of the absolute values of the differences between  points in the  MPs in Tables~\ref{tab:keystroke_anomaly_results}, \ref{tab:calf_anomaly_results} and \ref{tab:traffic_anomaly_results} though these are not normalised.

In experiments where we introduce duplicated anomalies x2, these duplicates appear at 5\% of the time thus should be equivalent to the introduction of irrelevant features at 5\%. However the resulting MPs are not equivalently dissimilar to the MP from the original data with large differences in  the results for the keystroke (9303 vs. 15333), calf movement (23585 vs 32825) and traffic (1739 vs. 2109) data sets (these results are bolded in Tables~\ref{tab:keystroke_anomaly_results}, \ref{tab:calf_anomaly_results} and \ref{tab:traffic_anomaly_results}  and in Table~\ref{tab:similarity_results} for convenience).
The reason for this is because the introduction of a duplicate  introduces a pattern, the duplicate itself which is  detected by the MP as a  motif whereas when introducing an irrelevant feature there is no pattern so its not a MP motif. 
This is shown by the max values in the traffic MP being higher as duplicates are added appearing as discords where the  maximum value in the MP for the original data goes from 2.25 to 4.09 when duplicated anomalies x2 are added. 

The calf data appears to be quite regular anyway so the  maximum value  of 8.28 in the original MP did not change much when duplicates and irrelevant noise features were added. The high minimum value of 0.63 in that MP indicates there were already some patterns in the calf data as calves have movement habits to do with their 24 h circadian rhythm. When either kind of  noise was introduced the patterns in the original data disappeared in the resulting MPs with the minimum value dropping from 0.63 for the original data to 0.06 and 0.05 when duplications x2 and irrelevant feature at 1\% were added.  As more noise was added, the noise itself had a pattern shown as the minimum values increased with more and more noise added.
For the keystrokes data in Table~\ref{tab:keystroke_anomaly_results} the minimum values were not affected by the addition of either kind of noise so MPs maintained their detection of patterns and the maximums remained approximately the same.

In an unsupervised setting where we do not have annotations or ground truth in the data to work with, we  cannot pre-filter anomalies unlike the approaches taken in \cite{10.1007/978-3-030-40014-9_5,icpram19} or in \cite{9792239}. Thus determining the capacity of the MP generation algorithm itslef to work with noisy data is important.
Our results have discovered that when the amount of noise in the original data is minor such as duplicates occurring at 5\% or irrelevant features at 1\%, the generated MPs are dissimilar to the MPs from the original data in approximate proportion to the amount of noist.  As more and more duplicates are added, the generated MPs get more and more dissimilar to the original MP but not to a significant extent and this is true for the keystrokes and calf data but not for the traffic data. An equivalent observation about adding irrelevant features cannot be made in Table~\ref{tab:similarity_results} because of the volume of noise added, up to 50\% at the extreme level, for any of the data sets.  For the traffic data, when an additional 50\% of irrelevant data is added, the dissimilarity value between the MP from the original and the noisy data, has risen to 1.434. The explanation for this is that the MP from the noisy data is so very different to the original MP that the DTW algorithm, even for a comparatively short time series of 3,600 values, cannot detect any equivalent pairs across the MPs.

\section{Conclusions}

The matrix profile has proved to be a useful tool in terms of making most time series data mining tasks intuitive and to require less effort compared to other methods such as using a dimensionality reduced representation via a brute force approach. The MP provides no false negatives or false positives in terms of motif and discord discovery, time series joins and classification via shapelet discovery for example, because it provides an exact solution. The ability to use the MP in a simple manner without any tuning, apart from selecting the window size, makes it relatively parameter free. 

This paper has presented three case studies where the matrix profile has enabled the identification of discords and motifs in relation to human typing behaviour, movement of new-born calves and  vehicular traffic flow in a city centre.
\textcolor{black}{The contribution of the paper is to demonstrate effect of introducing various forms of data corruption into the time series data and to show that such data corruption has} led to the generation of MPs which are similar to the MPs from the original data provided the amount of noise is small and only for some data sets.  \textcolor{black}{Our results have also shown that} once the amount of noise increases, the generated MPs are very different from the original MPs. If the original time series is short (N=3,600 in our case) then introducing even a small amount of noise leads to a more significant change in the generated MP. 

\textcolor{black}{The experimental results in this paper provide some direction to encourage further development of algorithms for generating the matrix profile where there is known to be different forms of data corruption including missing single and blocks of data, insertion of irrelevant data and transposition of one data value to another.}
Further work in this area should also examine the impact of noise on generating an incrementally updated matrix profile, on the impact of noise on hyper-large time series
and possibly on the introduction of other types of noise into the time series data.

\section*{Acknowledgements}

CH received no specific funding for this work.  The work of AS was part-funded  by Science Foundation Ireland under grant number SFI/12/RC/2289\_P2, co-funded by the European Regional Development Fund.  The funders had no role in study design, data collection and analysis, decision to publish, or preparation of the manuscript.

\bibliographystyle{plos2015}

%
%

%





\end{document}